\documentclass{IOS-Book-Article}

\usepackage{mathptmx}
\usepackage{graphicx}
\usepackage{url}
\usepackage{multirow}
\usepackage{tcolorbox}

\newtcolorbox{promptbox}[1][]{%
  colback=gray!10,
  colframe=gray!50,
  boxrule=0.5pt,
  arc=2pt,
  left=4pt,
  right=4pt,
  top=4pt,
  bottom=4pt,
  fontupper=\sffamily\tiny,
  width=\columnwidth,
  title=#1,
}
\begin{document}
\begin{frontmatter}              

\title{Knowledge Graph Completion for Action Prediction on Situational Graphs}
\subtitle{A Case Study on Household Tasks}

\author[A]{\fnms{Mariam} \snm{Arustashvili}\thanks{Corresponding Author: Mariam Arustashvili; University of Stavanger; E-mail: mariam.arustashvili@uis.no.}}, \author[B]{\fnms{Jörg} \snm{Deigmöller}} and \author[C]{\fnms{Heiko} \snm{Paulheim}}
\runningauthor{M. Arustashvili et al.}
\address[A]{University of Stavanger, Information Access and Artificial Intelligence group, Norway}
\address[B]{Honday Research Institute Europe, Germany}
\address[C]{University of Mannheim, Data and Web Science Group, Germany}

\begin{abstract}
Knowledge Graphs are used for various purposes, including business applications, biomedical analyses, or digital twins in industry 4.0. In this paper, we investigate knowledge graphs describing household actions, which are beneficial for controlling household robots and analyzing video footage. In the latter case, the information extracted from videos is notoriously incomplete, and completing the knowledge graph for enhancing the situational picture is essential. In this paper, we show that, while a standard link prediction problem, situational knowledge graphs have special characteristics that render many link prediction algorithms not fit for the job, and unable to outperform even simple baselines.
\end{abstract}

\end{frontmatter}

\thispagestyle{empty}
\pagestyle{empty}

\section{Introduction}

Knowledge graphs (KGs) are structured representations of knowledge that model entities and their relationships in a graph format, offering a semantically rich and machine-readable alternative to traditional databases \cite{ehrlinger2016defKG, hogan2021kgs}. By capturing complex relational data through nodes and edges, KGs enable advanced reasoning, inference, and data integration across heterogeneous sources \cite{nickel2016review}. Their versatility has made them essential in fields such as natural language processing, biomedical research, and personalized recommendation systems \cite{ji2022survey, nicholson2020biomed, wang2019kgat}. Specialized forms like situational and temporal knowledge graphs extend this functionality by incorporating dynamic or context-dependent information, supporting applications in rapidly changing environments such as robotics and context-aware systems \cite{cai2023temporal, kazemi2020replearn, krause2022generalized, zhang2021patterns}.

In robotics, KGs provide a structured framework for representing and reasoning about environments, enabling robots to interpret human actions, plan tasks, and adapt to new situations \cite{beetz2010towards}. Projects like RoboEarth \cite{waibel2011roboearth} and KnowRob \cite{tenorth2013knowrob} show how perception and action knowledge can be integrated to support autonomous behavior. Ontology-based approaches further highlight the value of structured reasoning for robotic perception, planning, and decision-making \cite{Alarcos2019}. Recent advances also show how multi-modal knowledge integration enhances embodied AI \cite{yaoxian2024scenedriven}.

A central challenge in robotics is reasoning under uncertainty. Robots operating in real-world environments—such as homes or workplaces—often face incomplete data due to occlusions, sensor limitations, or dynamic change \cite{patel2024robotic}. Knowledge Graph Completion (KGC) addresses this by inferring missing links (e.g., unobserved object interactions) or entities (e.g., likely tools) within structured graphs \cite{paulheim2016knowledge, rossi2021knowledge}.


Another challenge is learning patterns in action sequences. Robots must predict the next steps of human behavior based on context, rather than relying solely on commonsense assumptions or static rules. In this work, we investigate how well KGC methods support this task by testing them on situational KGs built from household activity videos. Specifically, we evaluate how different models predict a human’s goal (parent action) or next step (sub-action) in a sequence—e.g., predicting milk handling after observing cereal pouring—so the robot can assist appropriately.



We compare standard link prediction (LP) models (e.g., TransE, ComplEx), simple statistical baselines, and large language models (LLMs). Results show that conventional embedding models struggle to outperform even simple heuristics, likely due to the ambiguity and temporal nature of human activities. While LP remains central to KGC, our findings suggest that LLMs—thanks to their contextual reasoning and generalization abilities—may offer a more robust solution for parent action tasks. In contrast, sub-action prediction presents a greater challenge for LLMs and LP models, where simpler, pattern-based statistical methods outperform more complex approaches. We conclude with a discussion of open challenges and the potential of hybrid approaches.


\section{Related Work}

Using knowledge graphs to represent information about the surroundings is common in robotics~\cite {pistilli2023graph}. Semantic knowledge is essential for robots autonomy~\cite{Alarcos2019}, scene understanding in static and dynamic environments, and during human-machine interaction \cite{garg2020semantics}. Therefore, since graph-based data is used for knowledge representation in robots, it is common to make inferences such as action prediction on those situational graphs.

A prominent example of structured knowledge representation in robotics is the Functional Object-Oriented Network (FOON)~\cite{paulius2016functional}, which encodes task-related object and action knowledge using a bipartite graph structure. While our graph shares some structural similarities with FOON in terms of capturing object-action interactions, it differs in key ways: we emphasize temporal and hierarchical relations between sub-actions (e.g., "has\_next" and "has\_element") rather than focusing on functional units of manipulation. Additionally, our approach is data-driven, relying on per-frame annotations from the KIT Bimanual Actions Dataset, whereas FOON is often constructed manually or semi-manually.

Datasets that capture human activities with rich temporal and object-level detail are essential for advancing KGC and action prediction in robotics. Not many public datasets are available that allow experimentation with the action inference task using link prediction methods. Table~\ref{tab:datasets} shows the requirements that guided our dataset choice. The KIT Bimanual Actions Dataset~\cite{dreher2020learning} is a prominent example, originally developed for computer vision tasks and widely used in research on human-object interaction and bimanual action recognition. It consists of 540 RGB-D recordings across nine tasks performed by six subjects, with detailed, frame-level annotations of sub-actions for both hands. These annotations include spatial object data and allow for modeling fine-grained action sequences relevant to robotic planning and prediction.

\begin{table}
\begin{center}
\begin{tabular}{p{2.5cm} p{1cm} p{1.5cm} p{1.5cm} p{2cm} p{1.5cm}} 
 \hline
     Dataset & Parent actions & Sequential sub-actions & Time-stamps & Objects & States/ Properties\\ [0.5ex] 
     \hline\hline
     The Breakfast Actions Dataset \cite{Kuehne12} \cite{Kuehne16end} & Yes & Yes & Yes & Not directly, but possible to extract & No \\ 
     \hline
     \bfseries KIT Bimanual Actions Dataset \cite{dreher2020learning} & \bfseries Yes & \bfseries Yes & \bfseries Yes & \bfseries Yes & \bfseries Yes \\
     \hline
     AVA-Kinetics Dataset \cite{li2020avakinetics} & Yes & Yes & Yes & No & No \\
     \hline
     EPIC-KITCHENS-100 \cite{damen2022rescaling} & No & Yes & Yes & Yes & No \\
     \hline
     Georgia Tech Egocentric Activity Datasets \cite{li2020eyebeholder} & Yes & Yes & Yes & Yes & No \\
    \hline
     YouCook \cite{DaXuDoCVPR2013f} & Yes & Yes & Yes & Yes & No \\ [1ex] 
     \hline
\end{tabular}
\end{center}
\caption{Comparison of various action recognition datasets based on their features, including parent actions, sequential sub-actions, timestamps, objects, and states/properties.}
\label{tab:datasets}
\end{table}

\cite{zanchettin2018prediction} investigates predicting subsequent activities and estimating the time until an activity requiring assistance is initiated. This task is similar to what we do in this paper; however, they use a different methodology, namely Markov chains, to model human behavior and test it in a real-world setting. Literature is sparse on human action prediction using link prediction techniques. \cite{vesper2014support} discusses signaling in human-robot interaction as a tool to predict human actions. Other researchers perform activity prediction directly on visual data from videos \cite{kong2022human,liu2020forecasting}. In contrast, our work models the data derived from video input as a structured knowledge graph and explores link prediction methods on two tasks: (1) subsequent action prediction and (2) parent action prediction, i.e., predicting the overall task of an observed sequence of activities.


\section{Experiments}

This study leverages the KIT Bimanual Actions Dataset~\cite{dreher2020learning}, a structured benchmark originally designed for computer vision but well-suited for robotic action prediction due to its granular labeling of bimanual tasks. The dataset contains 540 recordings in which six subjects perform nine distinct tasks (parent actions)—such as preparing cereal or assembling tools—carried out in kitchen and workshop scenarios. Each task is repeated ten times and decomposed into sequential, fine-grained sub-actions (e.g., "approach bowl," "stir with whisk"). It provides frame-level ground truth for these sub-actions alongside 3D object interactions, enabling temporal and spatial modeling of action sequences. Key data is organized hierarchically in JSON files, with per-hand sub-action sequences, including start and end timestamps, and 3D bounding boxes for both hands and objects. During preprocessing, object–sub-action associations were established via bounding box intersection analysis. Temporal or spatial inconsistencies—such as missing or overlapping annotations—were manually corrected to ensure coherence.

\begin{figure}
    \centering
    \includegraphics[width=\linewidth]{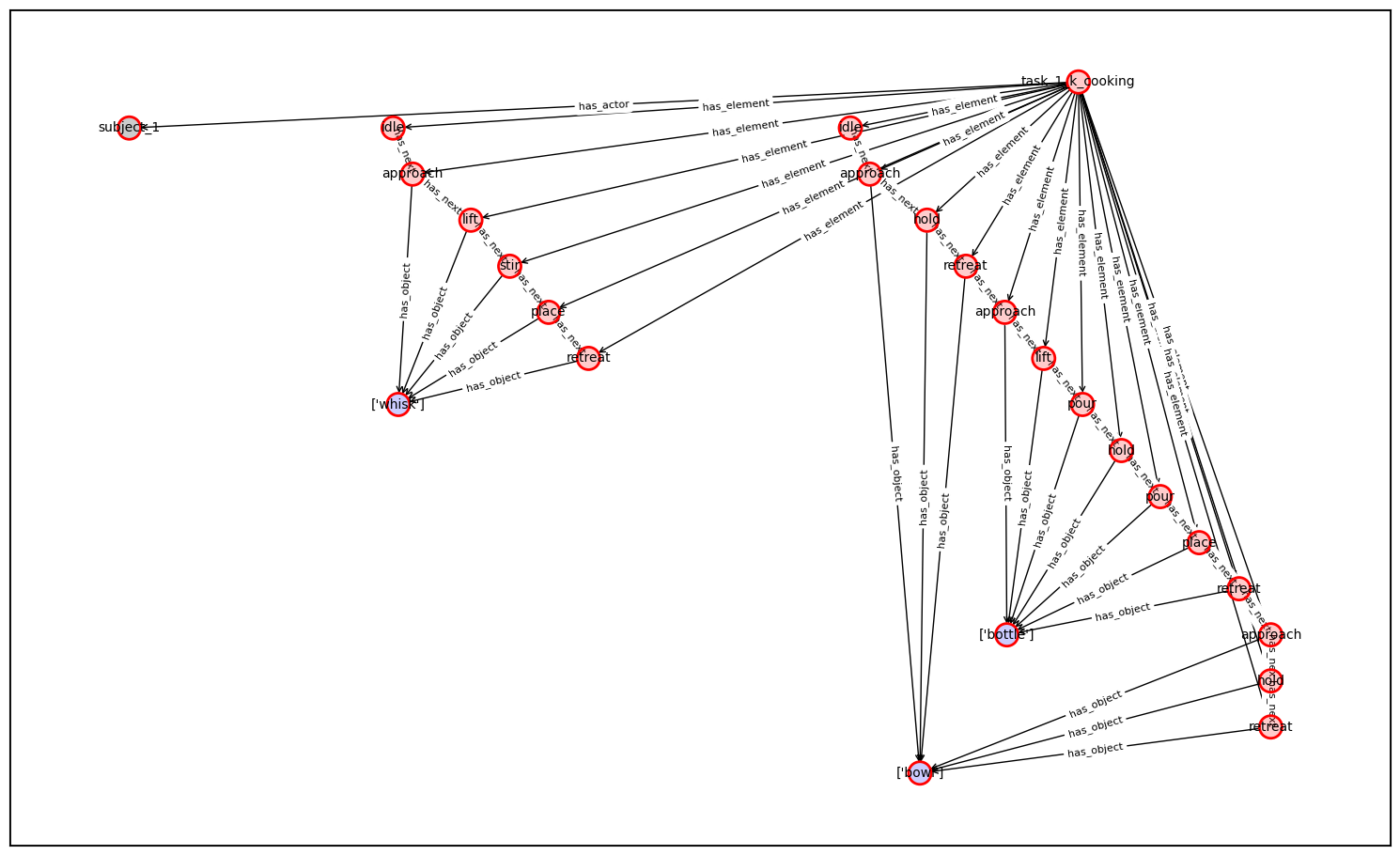}
    \caption{Sample graph visualization of the task cooking}
    \label{fig:graph_example}
\end{figure}

The graph structure was built using the following relation types: \textit{has\_actor} (link between parent action and the subject performing the task), \textit{has\_object} (link between sub-action and object), \textit{has\_element} (link between parent action and sub-action), and \textit{has\_next} (link between successive sub-actions). The refined data was structured to ensure compatibility with MemNet~\cite{memnet2023}, offering a dynamic and generalizable structure for encoding action–object–context relationships. The resulting KG contains 13,928 nodes and 32,577 edges; graph statistics are summarized in Table~\ref{tab:kgstats}. Each of the 540 task recordings forms a separate graph component, resulting in a fragmented and disconnected knowledge graph. Figure~\ref{fig:graph_example} illustrates a typical weakly connected component.

\begin{table}[h!] 
    \centering 
    \begin{tabular}{p{.65\textwidth}p{.3\textwidth}}
    \hline 
    \textbf{Statistic} & \textbf{Value} \\
    \hline
    \hline
     Number of Nodes & 13,928 \\
     Number of Edges & 32,577 \\
     Density & 0.00017 \\
     Average In-Degree & 2.34 \\
     Average Out-Degree & 2.34 \\ 
     Max In-Degree & 31 \\
     Max Out-Degree & 56 \\ 
     Number of Strongly Connected Components & 13,928 \\ 
     Number of Weakly Connected Components & 540 \\ 
     Average Clustering Coefficient & 0.289 \\ 
     Degree Assortativity Coefficient & -0.364 \\ 
     Reciprocity & 0.0 \\
    \hline
    \end{tabular} 
    \caption{Key Graph Statistics} 
    \label{tab:kgstats} 
\end{table}

For consistent and fair evaluation, we applied a fixed train–test split: two out of ten repetitions per task were reserved for testing. This yields 108 graph components in the test set, with 11,088 nodes and 25,897 edges in training, and 2,840 nodes and 6,680 edges in testing. This split is held constant across all models. The processed dataset, splits, and code are publicly available via the MemNet repository.\footnote{\url{https://github.com/HRI-EU/BasicMemnet/tree/master/data/action_sequences}}

We implemented baseline models for both parent action and sub-action prediction tasks to serve as performance benchmarks. These baselines are based on simple statistical assumptions. In both tasks, Baseline1 uses only action information, while Baseline2 incorporates contextual object information, providing a more nuanced view of the activity.

For the parent action prediction task, Baseline1 maps each sub-action to its most frequently co-occurring parent action in the training data. For each activity (i.e., a sequence of sub-actions), the final parent action is determined via majority vote across predicted labels. Baseline2 extends this by using pairs of sub-actions and their associated objects (e.g., "cut–knife", "hold–whisk") to build a more detailed mapping. Including object context significantly improves performance, as shown in Table~\ref{tab:results}.


For the subsequent action prediction task, Baseline1 predicts the most probable next sub-action based on the current one, using the most frequent transitions observed in training. Baseline2 considers the combination of the current sub-action and its associated object when determining the next step.

For knowledge graph completion, we evaluated several widely-used embedding-based link prediction models using the PyKeen framework~\cite{ali2021pykeen}: TransE~ \cite{bordes2013translating}, TransR~\cite{lin2015learning}, ComplEx~\cite{trouillon2016complex}, DistMult~\cite{yang2015embedding}, and RotatE~\cite{sun2019rotate}. Since action and sub-action entities are literal strings, we also tested LiteralE variants of DistMult and ComplEx\cite{kristiadi2019incorporating}, which incorporate textual literals into entity representations.

In addition to traditional KG models, we tested GPT-4o-mini~\cite{openai_gpt4omini} to evaluate how effectively a large language model can predict actions from structured KG data. To interface with the model, we transformed each graph instance into a set of textual triples describing actions, objects, and their relations. We then used few-shot prompting~\cite{brown2020language} with structured reasoning, where the model was shown several examples that included intermediate steps to guide its inference process (see Figure\ref{fig:prompt}). This setup allowed the model to learn patterns from analogous cases and apply them to new graph segments at inference time.
The code to replicate the results in this paper is available online.\footnote{\url{https://github.com/marusta/Knowledge_Refinement}}

\begin{figure}
    \centering
    \begin{promptbox}  
    You are a knowledgeable assistant. Given a set of triples representing sub-actions, predict the most likely parent action.
    \\
    \\
    Below are examples of parent actions and their corresponding sequences of sub-actions. Each triple represents a head, relation, and tail from the knowledge graph describing the parent action. 
    \\
    \\
    \{examples\}
    \\
    \\
    What is the parent action for the following graph?
    \\
    \\
    \{test set triples\}
    \\
    \\
    Answer nothing but one of the 9 possible parent actions: cooking, cooking\_with\_bowls, pouring, wiping, cereals, hard\_drive, free\_hard\_drive, hammering, sawing.
    \end{promptbox}
    \caption{
    Prompt used for parent action prediction. It follows a few-shot format with structured reasoning. Parent action labels are taken directly from the dataset, where some tasks are object-based rather than action-based. The same naming is used in the prompt examples.
    }
    \label{fig:prompt}
\end{figure}

\section{Results}

\begin{table}[t]
\centering
\begin{tabular}{llrrrrrr}
\hline
& & \multicolumn{3}{c}{\textbf{Parent Action Prediction}} & \multicolumn{3}{c}{\textbf{Subsequent Action Prediction}} \\
\hline
\textbf{Category} & \textbf{Model} & \textbf{Hits@1} & \textbf{Hits@3} & \textbf{Hits@5} 
& \textbf{Hits@1} & \textbf{Hits@3} & \textbf{Hits@5} \\
\hline
\hline
\textbf{Baselines} 
& Random & 11\% & 33\% & - 
& 7.14\% & - & - \\
& Baseline1 & 11\% & 76\% & - 
& 69.1\% & - & - \\
& Baseline2 & \textbf{76\%} & \textbf{100\%} & - 
& \textbf{81.7\%} & - & - \\
\hline
\textbf{KG Models} 
& TransE & 2.60\% & \textbf{4.90\%} & 5.30\% 
& 0.00\% & 53.40\% & \textbf{64.86\%} \\
& TransR & 3.25\% & 4.78\% & \textbf{5.72\%} 
& 0.00\% & 31.51\% & 45.10\% \\
& RotatE & \textbf{3.88\%} & 3.90\% & 3.92\% 
& \textbf{52.16\%} & \textbf{57.67\%} & 58.00\% \\
& ComplEx & 1.00\% & 2.00\% & 2.30\% 
& 8.56\% & 18.20\% & 24.18\% \\
& DistMult & 0.00\% & 0.00\% & 0.00\% 
& 33.68\% & 57.57\% & 58.84\% \\
& ComplExLiteral & 3.50\% & 3.80\% & 2.90\%
& 14.49\% & 27.14\% & 33.91\% \\ 
& DistMultLiteral & 1.50\% & 3.00\% & 3.40\% 
& 5.40\% & 11.38\% & 16.10\% \\ 
\hline
\textbf{Other} 
& GPT-4o-mini & \textbf{78.73\%} & - & - 
& 13.00\% & - & - \\
\hline
\end{tabular}
\caption{Parent action prediction and subsequent action prediction results}
\label{tab:results}
\end{table}

Our experiments reveal a clear performance hierarchy across methods. For parent action prediction, frequency-based baselines (76–100\% Hits@3) and GPT-4o-mini (78.73\% Hits@1) outperformed graph-based models ($\leq$ 5.72\% Hits@5), suggesting that statistical patterns or high-level reasoning suffice for coarse task recognition. In contrast, sub-action prediction favored heuristic baselines (81.7\% Hits@1), while graph models like RotatE (52.16\% Hits@1) showed limited but non-trivial capability—likely due to their ability to model localized action transitions. Strikingly, LLMs collapsed to 13\% accuracy here, exposing their weakness in fine-grained sequential reasoning.

This failure aligns with the previous studies \cite{ju2022IEEE} noting that many embedding-based models, such as TransE and TransR, struggle with multistep dependencies, which is crucial for accurate parent action prediction. It also highlights another limitation: disconnected subgraphs in robotic tasks violate the dense-connectivity assumptions of conventional KG benchmarks.

\section{Conclusion and Outlook}

This work identifies robotic action prediction as a critical testbed for KGC methods, where standard LP techniques underperform due to disconnected task subgraphs and hierarchical dependencies. While baselines and LLMs offer pragmatic solutions for specific tasks, their limitations underscore the need for specialized graph representations that encode temporal, compositional, and weakly connected structures.

Future research should redefine evaluation protocols for LP in disconnected graphs, moving beyond random link masking, develop hybrid architectures that integrate baseline robustness with KG relational reasoning, and explore dynamic graph embeddings to capture action progression.
By addressing these gaps, the field can unlock the KGC methods’ potential for real-world sequential decision-making.

\bibliographystyle{plain}
\bibliography{references}
\end{document}